\documentclass[10pt,twocolumn,letterpaper]{article}

\usepackage[pagenumbers]{cvpr}

\usepackage{amssymb} 
\usepackage{pifont}   
\newcommand{\greencheckmark}{\textcolor{green}{\checkmark}}
\newcommand{\redxmark}{\textcolor{red}{\ding{55}}}
\usepackage[table]{xcolor}

\usepackage{graphicx}
\usepackage{algorithm}
\usepackage{algorithmic}
\usepackage{amsmath}
\usepackage{booktabs}
\usepackage{amssymb}
\usepackage{mathtools}
\usepackage{multirow}

\definecolor{cvprblue}{rgb}{0.21,0.49,0.74}
\usepackage[pagebackref,breaklinks,colorlinks,allcolors=cvprblue]{hyperref}

\title{Understanding the Implicit User Intention via Reasoning with Large Language Model for Image Editing}

\author{Yijia Wang$^1$ \quad Yiqing Shen$^2$ \quad Weiming Chen$^1$ \quad Zhihai He$^{1,3}$\thanks{Corresponding author}\\
$^1$Southern University of Science and Technology, Shenzhen, China\\
$^2$Johns Hopkins University, Baltimore, USA\\
$^3$Pengcheng Laboratory, Shenzhen, China\\
{\tt\small \{wangyj2022,chenwm2023\}@mail.sustech.edu.cn, yiqingshen1@gmail.com}\\
{\tt\small hezh@sustech.edu.cn}
}

\begin{document}
\maketitle

\begin{abstract}
Existing image editing methods can handle simple editing instructions very well. To deal with complex editing instructions, 
they often need to jointly fine-tune the large language models (LLMs) and diffusion models (DMs), which involves very high computational complexity and training cost. 
To address this issue, we propose a new method, called  \textbf{C}omplex \textbf{I}mage \textbf{E}diting via \textbf{L}LM \textbf{R}easoning (CIELR), which converts a complex user instruction into a set of simple and explicit editing actions, eliminating the need for jointly fine-tuning the large language models and diffusion models. 
Specifically, we first construct a structured semantic representation of the input image using foundation models. 
Then, we introduce an iterative update mechanism that can progressively refine this representation, obtaining a fine-grained visual representation of the image scene. This allows us to perform complex and flexible image editing tasks.
Extensive experiments on the SmartEdit Reasoning Scenario Set show that our method surpasses the previous state-of-the-art by 9.955 dB in PSNR, indicating its superior preservation of regions that should remain consistent.
Due to the limited number of samples of public datasets of complex image editing with reasoning, we construct a benchmark named CIEBench, containing 86 image samples, together with a metric specifically for reasoning-based image editing. CIELR also outperforms previous methods on this benchmark.
The code and dataset are available at \href{https://github.com/Jia-shao/Reasoning-Editing}{https://github.com/Jia-shao/Reasoning-Editing}.
\end{abstract}

\section{Introduction}
\label{sec:intro}

Image editing aims to modify specific regions of an image according to user instructions while preserving consistency across unmodified areas \cite{perarnau2016invertible, cheng2020sequential}. 
Diffusion models (DMs) have demonstrated their capabilities and applications in image editing \cite{brooks2023instructpix2pix, geng2024instructdiffusion, zhang2023magicbrush, hertz2022prompt, mokady2023null, feng2024ranni}. 
However, existing DM-based image editing methods usually rely on manually defined masks to specify regions for modification, which demands considerable user expertise, and proves time-consuming, as shown in Tab.~\ref{tab:compare_feature}. 
To enable more straightforward iteration, text-prompted editing methods such as InstructPix2Pix \cite{brooks2023instructpix2pix}, Prompt-to-Prompt \cite{hertz2022prompt}, DiffEdit \cite{couairon2022diffedit}, and Null-text Inversion \cite{mokady2023null} have been proposed to enable users to perform image editing via natural language text queries without manual masks.
Nevertheless, these approaches struggle with understanding the user's intention to perform complex editing tasks, which often require reasoning based on world knowledge,  \eg ``\textit{replace the food containing the most vitamin C with an orange}''.

To handle complex image editing tasks, recent methods \cite{huang2024smartedit, yu2024anyedit, jin2024reasonpix2pix} rely on Large Language Models (LLMs) \cite{touvron2023llama, chiang2023vicuna} to interpret more complex queries for image editing. These methods suffer from two major challenges.
First, they typically require joint fine-tuning of both LLMs and DMs, which is very costly.
For example, SmartEdit \cite{huang2024smartedit} utilizes LoRA \cite{hu2022lora} to fine-tune its LLM component with pipeline-wide updates during each model iteration. 
Second, these methods often struggle with complex queries that require multi-step reasoning, as they lack an interpretable intermediate representation that can be iteratively refined to break down complex reasoning tasks into manageable steps.
This limitation becomes particularly pronounced in multi-step editing scenarios where previous edits affect subsequent reasoning and editing operations.

\begin{table}[t!]
\centering
\caption{
Comparative analysis of image editing methods across main capabilities.
\textbf{DMA} represents ``diffusion model agonistic''; \textbf{Reas} denotes reasoning capability for complex implicit queries; \textbf{Auto} represents fully automatic operation without manual interaction; \textbf{RM.} (Remove), and \textbf{RP.} (Replace) indicate supported editing operations; and \textbf{MS.} represents multi-step editing capability for complex reasoning scenarios.
Our CIELR (highlighted) uniquely supports all capabilities, especially multi-step reasoning through its chain of structured semantic representation updates.
}\label{tab:compare_feature}
\resizebox{\linewidth}{!}{
\begin{tabular}{l|ccccccc}
\hline 
\textbf{Method} & \textbf{DMA} & \textbf{Reas} & \textbf{Auto} & \textbf{Add} & \textbf{RM.} & \textbf{RP.} & \textbf{MS.} \\
\hline 
\rowcolor[gray]{0.95}
\textbf{BrushEdit} \cite{li2024brushedit}& \greencheckmark & \redxmark & \redxmark & \greencheckmark & \greencheckmark & \greencheckmark & \redxmark \\
\textbf{ACE++} \cite{mao2025ace++}& \redxmark & \redxmark & \redxmark & $\greencheckmark$ & \greencheckmark & \greencheckmark & \redxmark \\
\rowcolor[gray]{0.95}
\textbf{AnyEdit} \cite{yu2024anyedit} & \greencheckmark & \greencheckmark & \greencheckmark & $\greencheckmark$ & $\greencheckmark$ & $\greencheckmark$ & \redxmark \\
\textbf{Add-it} \cite{tewel2024add}& \redxmark & \redxmark & \greencheckmark & $\greencheckmark$ & \redxmark & \redxmark & \redxmark \\
\rowcolor[gray]{0.95}
\textbf{SmartEdit} \cite{huang2024smartedit}& \redxmark & \greencheckmark & \greencheckmark & \redxmark & $\greencheckmark$ & \greencheckmark & \redxmark \\
\textbf{InstructPix2Pix} \cite{brooks2023instructpix2pix}& \redxmark & \redxmark & \greencheckmark & $\greencheckmark$ & \greencheckmark & \greencheckmark & \redxmark \\
\rowcolor[HTML]{FFEBEA}
\hline 
\textbf{CIELR} & \greencheckmark & \greencheckmark & $\greencheckmark$ & $\greencheckmark$ & \greencheckmark & \greencheckmark & \greencheckmark \\
\hline
\end{tabular}
}
\end{table}

To address these limitations, we propose Complex Image Editing via LLM Reasoning (CIELR), a new reasoning-based editing framework that decouples reasoning from editing using the structured semantic representation of the input image as an interpretable intermediate layer, as shown in \cref{fig:overview}.
Our structured semantic representation consists of spatial (segmentation masks), semantic (object labels), and depth information extracted from the image, providing LLMs with a comprehensive yet abstract understanding of visual content while filtering irrelevant details that might impede reasoning.
A major innovation of our approach is that, unlike previous methods, CIELR eliminates the need for joint fine-tuning of LLMs and DMs, as both the structured semantic representation construction pipeline and reasoning module operate in a zero-shot manner. This makes our framework both LLM-agnostic and DM-agnostic, enabling seamless integration of state-of-the-art foundation models without costly retraining.
When information gaps are identified during reasoning, the system dynamically updates the structured semantic representation with additional spatial or semantic details until it contains sufficient information to complete the reasoning process.
The final reasoning output, comprising precisely identified target regions and explicit editing instructions,  directs the diffusion model to execute desired modifications while maintaining consistency in unmodified regions.

\section{Related Work and Unique Contributions}

In this section, we review existing training-free image editing methods related to our work and summarize our unique contributions.

\subsection{Training-free Image Editing with Diffusion Models}
\label{sec: training-free image editing methods}
In recent years, diffusion models \cite{dhariwal2021diffusion, saharia2022photorealistic, Esser2024RectifiedFlowTransformers, Chen2024PixArt-alpha, Sauer2024SDXLTurbo, Podell2024SDXL, Betker2023DALL-E3, Balaji2023eDiff-I} have significantly assisted the image editing task. Lots of methods achieve this without training. For example, Prompt-to-Prompt \cite{hertz2022prompt} transfers the attention maps generated during source-prompt inversion into the denoising process guided by the target prompt, thereby enabling precise semantic image editing. MasaCtrl \cite{Cao2023MasaCtrl} replaces the bidirectional self-attention key and value in the diffusion denoising process. DragonDiffusion \cite{mou2023dragondiffusion} uses classifiers to guide and cross-branch self-attention, translating users' drag or mask signals into geometric and texture offsets in latent space. DiffUHaul \cite{avrahami2024diffuhaul} uses mask gating for self-attention and soft anchoring constraints, enabling the target object to move along the user's trajectory without disturbing the background details. Eedit \cite{yan2025eedit} caches and reuses key features during the inversion phase, and dynamically skips redundant calculations with regional score rewards to achieve one-click real-time editing. KV-Edit \cite{zhu2025kv} reuses DiT's KV cache to freeze background representations and only perform denoising updates on specified areas to keep the consistency of areas that don't require editing. However, these methods exhibit satisfactory performance only when the editing instructions are straightforward; they falter when the instructions necessitate compositional or logical reasoning.

\begin{figure}
    \centering
    \includegraphics[width=\linewidth]{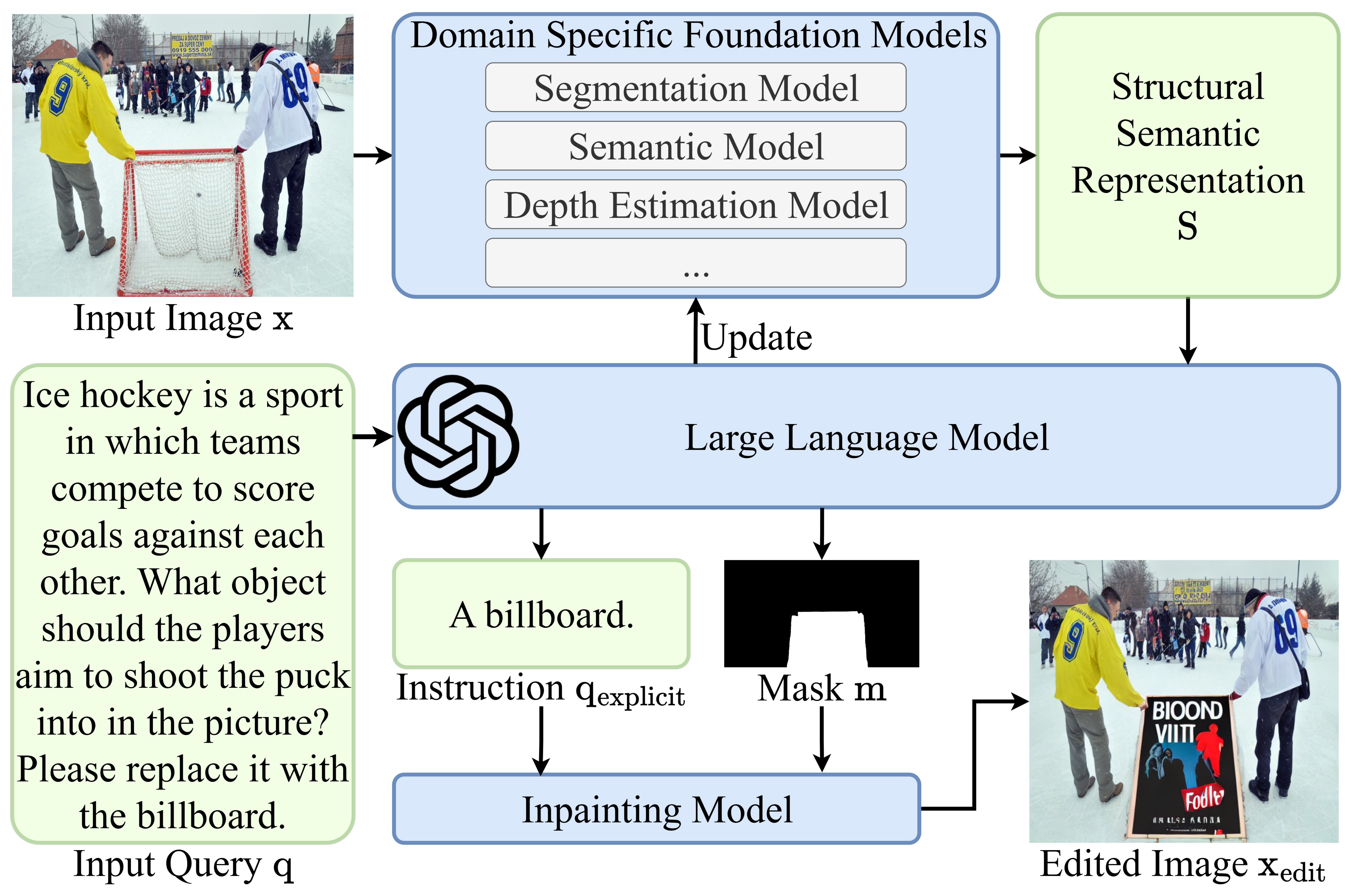}
    \caption{Overview of our proposed CIELR framework. The input is an image and a complex implicit editing instruction, and the output is an edited image.} 
    \label{fig:overview}
\end{figure}

\subsection{Reasoning-Based Visual Tasks}
\label{sec: reasoning visual tasks}

Significant progress has been made in the research of reasoning visual tasks, with a focus on multimodal learning and reasoning, multi-step dynamic reasoning, visual relationship detection, visual commonsense reasoning, and the exploration of new paradigms \cite{zhang2024llama, xu2024llava, zhang2024critic}. For instance, the modular neural network model \cite{ilievski2017multimodal} significantly enhances the model's reasoning ability by integrating visual and linguistic information. The Chain of Reasoning (CoR) model \cite{wu2018chain} supports multi-step dynamic reasoning. Additionally, new tasks such as Transformation Driven Visual Reasoning (TVR) \cite{hong2021transformation} have been proposed. However, they fail in generating images. Therefore, some methods \cite{koh2023generating, ge2307planting, ge2023making} try to address this gap by serving as a bridge between LLMs and DMs. Although the reasoning and generalization abilities of models have been significantly improved, they can't work in the image reasoning editing task.
Ranni \cite{feng2023ranni} uses LLMs to first break down complex text prompts into editable structured semantic panels, and then has the diffusion model generate them, but it works poorly on complex editing instructions that need reasoning. SmartEdit \cite{huang2024smartedit} introduces a bidirectional interaction module (BIM) to enhance the integration of MLLMs with diffusion models, thereby improving the model's ability to comprehend and execute complex instructions. Still, they struggle in the joint fine-tuning of both LLMs and DMs.

\subsection{Unique Contributions}
\label{sec: unique contributions}
Compared to existing work, our unique contributions can be summarized as follows: (1) We develop a novel complex image editing framework, called \textit{CIELR}, which uses structured semantic representations as an intermediate interpretable layer to decouple reasoning from editing, eliminating the need for joint fine-tuning of LLMs and DMs. (2) We design a chain of structured semantic update methods that enable multi-step reasoning through progressive refinement of structured semantic representations, addressing complex implicit queries that require world knowledge and inference. (3) We construct CIEBench, a benchmark dataset containing 86 image-query-mask triplets specifically designed to test reasoning-based editing capabilities, along with the Image Difference Check Score (IDCS), a metric specifically for the reasoning-based image editing task that evaluates semantic correctness rather than just visual similarity. (4) Extensive experiments on three datasets demonstrate that our method achieves superior performance on image editing with both implicit and explicit queries.

\section{Methods}
\label{sec: methods}

In this section, we provide the definition of the problem (\cref{sec: problem definition}) and an overview of our method (\cref{sec: method overview}), and then describe the process of constructing and updating structured semantic representations of images (\cref{sec: structured semantic representation construction and update}). 

\subsection{Problem Definition}
\label{sec: problem definition}

We first provide a formal definition for the task of image reasoning editing (RE), which requires edit operations based on implicit text queries demanding multi-step reasoning.
Unlike conventional image editing tasks that rely on explicit text query (\eg, ``\textit{replace the apple with an orange}''), RE addresses complex implicit text queries requiring inference and world knowledge (\eg, ``\textit{replace the food containing the most vitamin C with an orange}'').
It, therefore, necessitates an understanding of both visual content and semantic relationships, similar to reasoning segmentation \cite{lai2024lisa} but with the additional challenge of making appropriate edits while keeping the other parts consistent.
Formally, given an input image $\mathrm{x}$ and an implicit text query $\mathrm{q}$, the objective of RE is to generate an edited image $\mathrm{x}_\text{edit}$ that satisfies $\mathrm{q}$ without additional interaction, formulated as $\mathrm{x}_\text{edit} = \mathcal{F}_\text{RE}(\mathrm{x}, \mathrm{q})$, where $\mathcal{F}_\text{RE}$ represents the RE model that maps the input image and query to the edited output.

We categorize RE into three types. 
The first type, namely source object identification, requires identifying and locating the objects in the image needed to be edited, depicted as $\mathrm{x}_\text{edit} = \mathcal{F}_\text{edit}(\mathrm{x}, \mathrm{q}, \mathcal{F}_\text{id}(\mathrm{x}, \mathrm{q}))$ where $\mathcal{F}_\text{id}$ is the reasoning model that identifies the source object.
The editing model $\mathcal{F}_\text{edit}(\mathrm{x},\mathrm{q},\mathrm{m})$ takes the input image $\mathrm{x}$, region identifying the source object $\mathrm{m}$, and text query $\mathrm{q}$. 
The second type is to identify the targeted objects to be edited into, represented as $\mathrm{x}_\text{edit} = \mathcal{F}_\text{edit}(\mathrm{x}, \mathcal{F}_\text{tgt}(\mathrm{x}, \mathrm{q}),\mathrm{m})$, where $\mathcal{F}_\text{tgt}$ is the model that determines the appropriate target object.
The third type, multi-step edits, combines multiple reasoning-based modifications in sequence, which can be represented as a composition of operations
$\mathrm{x}_\text{edit} = \mathcal{F}_\text{RE}^{(n)} \circ \mathcal{F}_\text{RE}^{(n-1)} \circ ... \circ \mathcal{F}_\text{RE}^{(1)}(\mathrm{x}, \mathrm{q})$, where each $\mathcal{F}_\text{RE}^{(j)}$ ($j \in \{1, 2, ..., n\}$) represents an individual reasoning editing operation. 
Typically, one real-world RE task can involve one or multiple types; therefore, we will not differentiate them in the remaining context. 

\begin{figure}
    \centering
    \includegraphics[width=\linewidth]{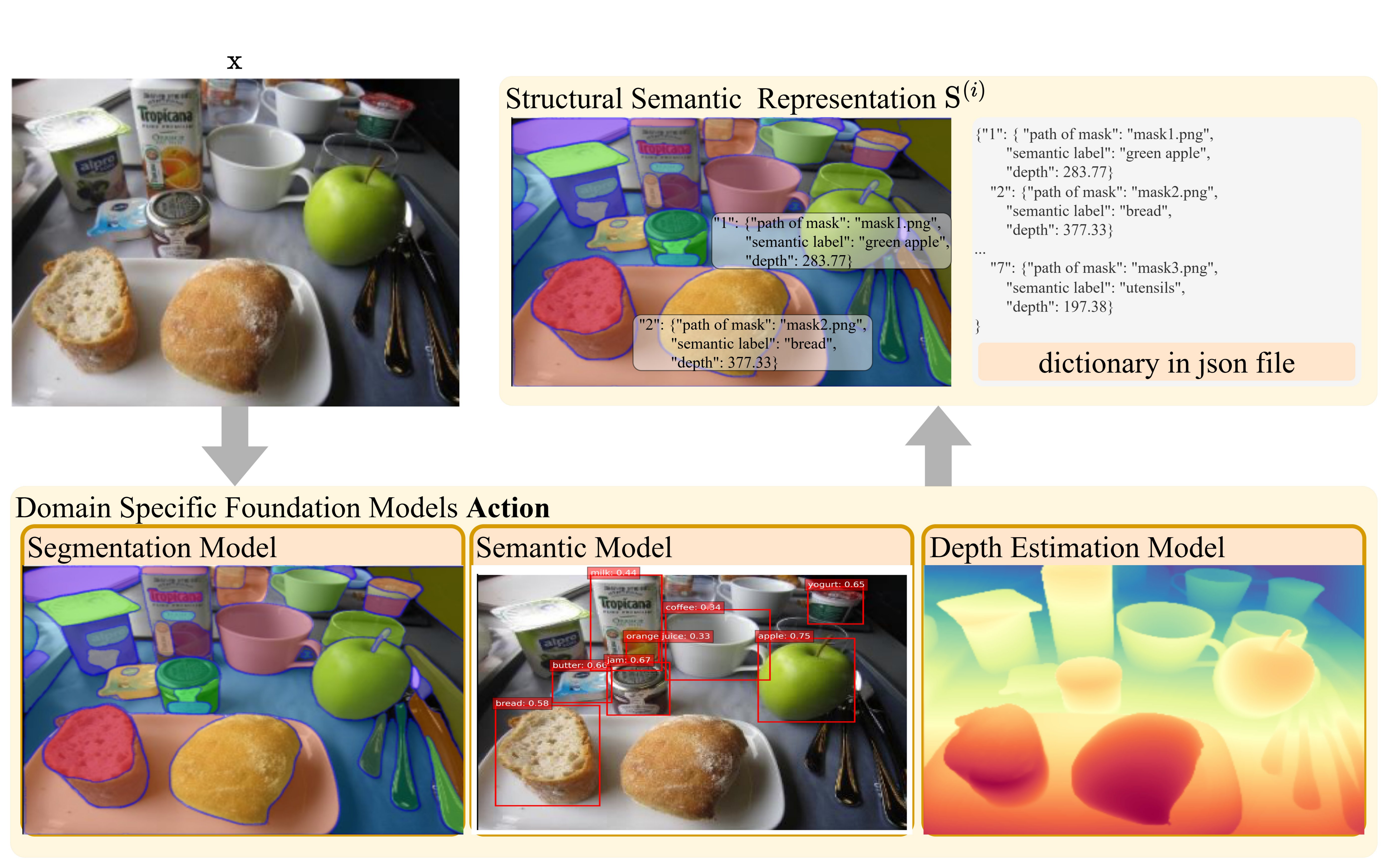}
    \caption{Detailed illustration of constructing the structured semantic representation and the final structured semantic representation in dictionary format.}
    \label{fig:DT_construct}
\end{figure}

\subsection{Method Overview}
\label{sec: method overview}
As depicted in \cref{fig:overview}, our method consists of three components, namely the initialized structured semantic representation construction, a chain of structured semantic representation updates for multi-step reasoning, and image editing execution by inpainting models.
Initially, we construct a structured semantic representation $\mathrm{S}^{(0)}$ by processing the original image $\mathrm{x}$ through a fixed structured semantic representation construction pipeline $\mathcal{F}_{\text{ssr}}$, represented as $\mathrm{S}^{(0)} = \mathcal{F}_{\text{ssr}}(\mathrm{x}, \mathrm{q})$. 
This $\mathcal{F}_{\text{ssr}}$ integrates domain-specific foundation models for spatial and semantic understanding to create an initialized intermediate representation of the image content. 
Then, in the chain of structured semantic representation update stage, when information gaps are identified, we update the structured semantic representation initialized by $\mathrm{S}^{(0)}$ through an iterative refinement process, depicted as $\mathrm{S}^{(i)} = \mathcal{F}_{\text{ssr}} (\mathrm{S}^{(i-1)},\mathrm{q})$.
We conduct this iteration until the structured semantic representation contains sufficient information for an LLM $\mathcal{F}_{\text{llm}}$ to identify the targeted region and the explicit query to perform editing, namely $\mathrm{m}, \mathrm{q}_{\text{explicit}}= \mathcal{F}_{\text{llm}}(\mathrm{S}^{(i)},\mathrm{q})$, where $\mathrm{m}$ represents the binary mask to identify the target editing region and $\mathrm{q}_{\text{explicit}}$ denotes the simplified explicit instruction.
Finally, we execute the image editing using a DM $\mathcal{F}_{\text{edit}}$, denoted as $ \mathrm{x}_{\text{edit}} = \mathcal{F}_{\text{edit}}
(\mathrm{x},\mathrm{q}_{\text{explicit}},\mathrm{m})$.
For multi-step editing scenarios, each edited image becomes the input for subsequent editing steps, enabling complex sequential transformations without requiring specialized multi-step editing models, namely
\begin{equation}
    \mathrm{x}_{\text{edit}}^{(j)} = \mathcal{F}_{\text{edit}}(\mathrm{x}_{\text{edit}}^{(j-1)},\mathrm{q}_{\text{explicit}}^{(j)},\mathrm{m}^{(j)}),
\end{equation}
where $j$ indicates the editing step.

\subsection{Structured Semantic Representation}
\label{sec: structured semantic representation construction and update}

In this section, we present our method to construct and update the structured semantic representation of the input image.
The structured semantic representation $\mathrm{S}^{(i)}$ is always structured as a dictionary that aims to preserve the spatial, semantic, depth, and any other necessary information of objects within the image to conduct the reasoning by a LLM.
As illustrated in \cref{fig:DT_construct}, the initial structured semantic representation $\mathrm{S}^{(0)}$, each object in the image is assigned a unique identifier $\mathcal{ID}$ and associated with three attributes: $\mathcal{P}$ (path to binary mask to identify the precise region of the object), $\mathcal{SL}$ (semantic label for this object), and $\mathcal{D}$ (relative depth). 
The construction of $\mathrm{S}^{(0)}$ leverages three specialized foundation models in a sequential pipeline. 
First, we employ a segmentation foundation model (\ie SAM2 \cite{ravi2024sam}) to decompose the image into distinct instance regions represented by binary masks $\mathcal{P}$. 
Next, a semantic foundation model (\ie OWLv2 \cite{minderer2024scaling}) extracts the semantic information by giving a bounding box and the corresponding $\mathcal{SL}$.
To establish correspondence between the segmented regions and semantic labels tagged on a bounding box, we calculate the IoU between each binary mask and the bounding boxes, where the semantic label associated with each bounding box yielding the highest IoU is assigned to the corresponding binary mask. 
Since segmentation models often produce overly fine-granular results, we implement a merging strategy where masks with identical semantic labels are consolidated into unified objects. 
Specifically, we identify adjacent or overlapping masks sharing identical semantic labels and combine their binary masks through a union operation, treating them as a single semantic entity with a unified contour.
Finally, a depth foundation model (\ie DepthAnything \cite{yang2024depth}) provides depth information $\mathcal{D}$ for each instance. 
For each identified object, we compute the median depth value across all pixels within its binary mask, which is then assigned as the representative depth for the entire object.

\begin{figure}[t]
    \centering
    \includegraphics[width=1.0\linewidth]{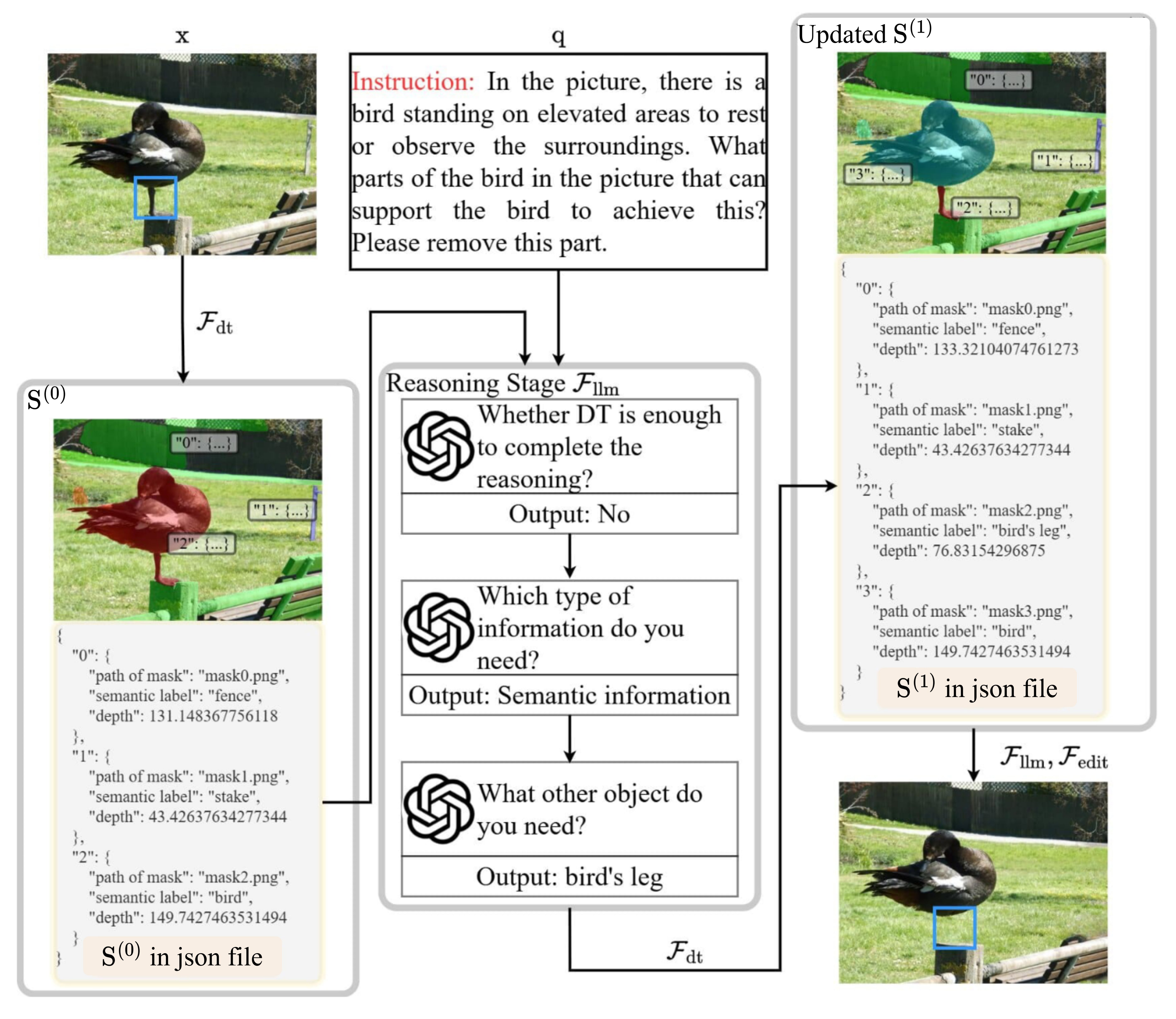}
    \caption{
    Detailed illustration of the chain of structured semantic representation update process on a sample from CIEBench. 
    The left panel shows the initial structured semantic representation $\mathrm{S}^{(0)}$ 
    The right panel presents the updated structured semantic representation $\mathrm{S}^{(1)}$ with newly added semantic details, enabling the LLM to successfully identify the region requiring modification, resulting in the precisely edited output image (bottom right).
    }
    \label{fig:chain}
\end{figure}

\subsection{Iterative Update of the Structured Semantic Representation}
Inspired by CoT \cite{wei2022chain}, our chain of structured semantic representation update method aims to enable multi-step reasoning for complex implicit editing queries. 
To be more specific, it progressively refines the structured semantic representation $\mathrm{S}^{(i)}$ until it contains sufficient information for the LLM to determine precise editing instructions and objects to be edited.
As shown in \cref{fig:chain}, the chain of structured semantic update operates through three sequential phases.
Firstly, given an input image $\mathrm{x}$ and an implicit text query $\mathrm{q}$, we prompt LLM to analyze the complexity of the editing task by $\mathcal{C} = \mathcal{F}_{\text{llm}}^{\text{complexity}}(\mathrm{q})$, where $\mathcal{C}$ represents the set of sub-tasks of RE that decompose the initial implicit text query.
For single-step edits, $\mathcal{C} = \{\mathrm{q}\}$; for multi-step edits, $\mathcal{C} = \{\mathrm{q}_1, \mathrm{q}_2, ..., \mathrm{q}_n\}$ where $n$ is number of steps, each $\mathrm{q}_j$ ($j \in \{1, 2, ..., n\}$) represents a decomposed sub-task from $\mathrm{q}$.
Second, we assess whether the current structured semantic representation $\mathrm{S}^{(i)}$ contains sufficient information to complete the reasoning required for each sub-task, formulated as
\begin{equation}
    \mathrm{I}_j = \mathcal{F}_{\text{llm}}^{\text{assess}}(\mathrm{S}^{(i)}, \mathrm{q}_j), \quad \text{For}~j \in \{1, 2, ..., n\},
\end{equation}
where $\mathrm{I}_j \in \{ \text{Spatial},\text{Semantic}\}$ specifies what type of additional information is needed; otherwise, $\mathrm{I}_j=\emptyset$ if the current structured semantic representation is adequate for LLM reasoning.
When $\mathrm{I}_j\not=\emptyset$, we update the structured semantic representation $\mathrm{S}^{(i)}$ based on the type of missing information $\mathrm{I}_j$, namely
\begin{equation}
\mathrm{S}^{(i+1)} \hspace{-4pt} = \hspace{-4pt}
\begin{cases}
\mathcal{F}_{\text{semantic}}(\mathrm{x}, \mathrm{q}_j, \mathrm{I}_j),
& \hspace{-10pt} \mathrm{I}_j \hspace{-3pt} = \hspace{-3pt} \text{Semantic}  \\
\mathcal{F}_{\text{spatial}}(\mathrm{S}^{(i)}, \mathcal{F}_{\text{llm}}^{\text{code}}(\mathrm{S}^{(i)}, \mathrm{I}_j)),
& \hspace{-10pt} \mathrm{I}_j \hspace{-3pt} = \hspace{-3pt} \text{Spatial} \\
\end{cases}
\end{equation}
where $\mathcal{F}_{\text{semantic}}$ represents the structured semantic representation update operation with additional semantic requirements.
Specifically, $\mathcal{F}_{\text{sematic}}$ refines the existing structured semantic representation by another LLM that re-analyzes the $\mathrm{S}^{(i)},\mathrm{q}, \mathrm{x}$ with targeted focus on the specific semantic categories identified in $\mathrm{I}_j$, adjusting segmentation thresholds in SAM2 \cite{ravi2024sam} and prompting OWLv2 \cite{minderer2024scaling} to identify previously undetected objects. 
The operation $\mathcal{F}_{\text{spatial}}$ executes LLM-generated Python code $\mathcal{F}_{\text{llm}}^{\text{code}}$ to extract additional spatial relationships, relative positions, or geometric properties that enrich the structured semantic representation beyond the basic attributes ($\mathcal{P}$, $\mathcal{SL}$, $\mathcal{D}$).
This update phase iterates until the structured semantic representation becomes sufficient, namely $\mathrm{I}_j = \emptyset$.
Finally, once a sufficiently structured semantic representation is achieved (denoted as $\mathrm{S}^{(I)}$), the LLM performs reasoning to identify the specific objects referenced by the implicit query and generates an explicit editing instruction:
\begin{equation}
    \mathrm{m}_j, \mathrm{q}_{\text{explicit},j} = \mathcal{F}_{\text{llm}}^{\text{reason}}(\mathrm{S}^{(I)}, \mathrm{q}_j),
    \label{eq:mask}
\end{equation}
where $\mathrm{m}_j$ represents the binary mask identifying the region to be edited with respect to $\mathrm{q}_j$, and $\mathrm{q}_{\text{explicit},j}$ represents the simplified explicit instruction derived from the implicit query $\mathrm{q}_j$.

\subsection{Reasoning-Based Image Editing}
The final component in CIELR is to execute the image editing operations based on the outputs from the chain of structured semantic representation update. 
After obtaining the binary mask $\mathrm{m}_j$ identifying the region to be edited and the explicit instruction $\mathrm{q}_{\text{explicit},j}$ from Eq.~\eqref{eq:mask}, we leverage a diffusion model to perform the actual image editing.
For each editing sub-task $\mathrm{q}_j \in \mathcal{C}$, we execute the image editing operation as follows:
\begin{equation}
    \mathrm{x}_{\text{edit},j}\hspace{-3pt} = \hspace{-3pt}\mathcal{F}_{\text{edit}}(\mathrm{x}_{\text{input},j}, \mathrm{q}_{\text{explicit},j}, \mathrm{m}_j),  ~j  \in \{1, 2, ..., n\},
\end{equation}
where $\mathrm{x}_{\text{input},j}$ represents the input image for the current editing step (either the original image $\mathrm{x}$ for the first sub-task or the result from the previous editing step) and $\mathcal{F}_{\text{edit}}$ is the diffusion model-based editing model.
The $\mathcal{F}_{\text{edit}}$ can be implemented using any contemporary diffusion model that supports region-based editing operations. 
Note that our framework is, therefore, diffusion-model-agnostic in this regard, allowing flexible integration with various state-of-the-art diffusion models without requiring any model fine-tuning, which aligns with our goal of decoupling reasoning from editing.
For multi-step editing scenarios, we apply the editing operations sequentially.

\begin{figure}[t]
    \centering
    \includegraphics[width=\linewidth]{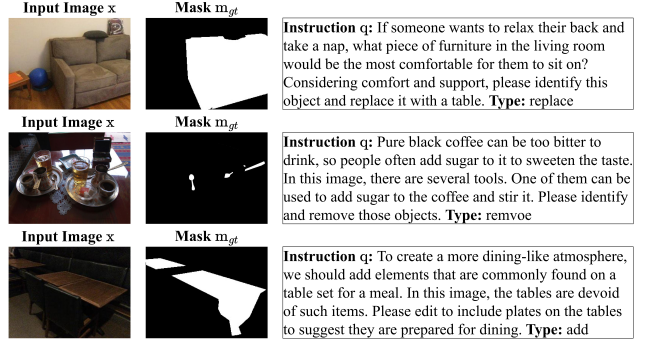}
    \caption{
    Representative samples from our CIEBench dataset showing the three types of reasoning editing tasks. 
    }
    \label{fig:dataset}
\end{figure}

\section{A New Benchmark Dataset}
\label{sec: proposed benchmark and evaluation metric}
In this paper, for a more comprehensive evaluation of reasoning-based image editing methods, we introduce a new benchmark dataset that contains a much larger number of test images with more complex editing tasks.

Existing datasets primarily focus on straightforward manipulations with explicit instructions. Hence, we construct CIEBench by using the images and masks from the reasoning segmentation dataset ReasonSeg \cite{lai2024lisa}.
For CIEBench, we carefully selected 86 high-quality images from ReasonSeg \cite{lai2024lisa} and manually transformed the original segmentation instructions into implicit, reasoning-intensive editing queries. 
Specifically, each data sample in CIEBench comprises an input image $\mathrm{x}$ to be edited, an implicit reasoning query $\mathrm{q}$ requiring multi-step inference, and a binary mask $\mathrm{m}_{gt}$ indicating ground truth regions to be edited, as shown in Fig.~\ref{fig:dataset}. 

\begin{figure*}[t!]
    \centering
    \includegraphics[width=0.9\linewidth]{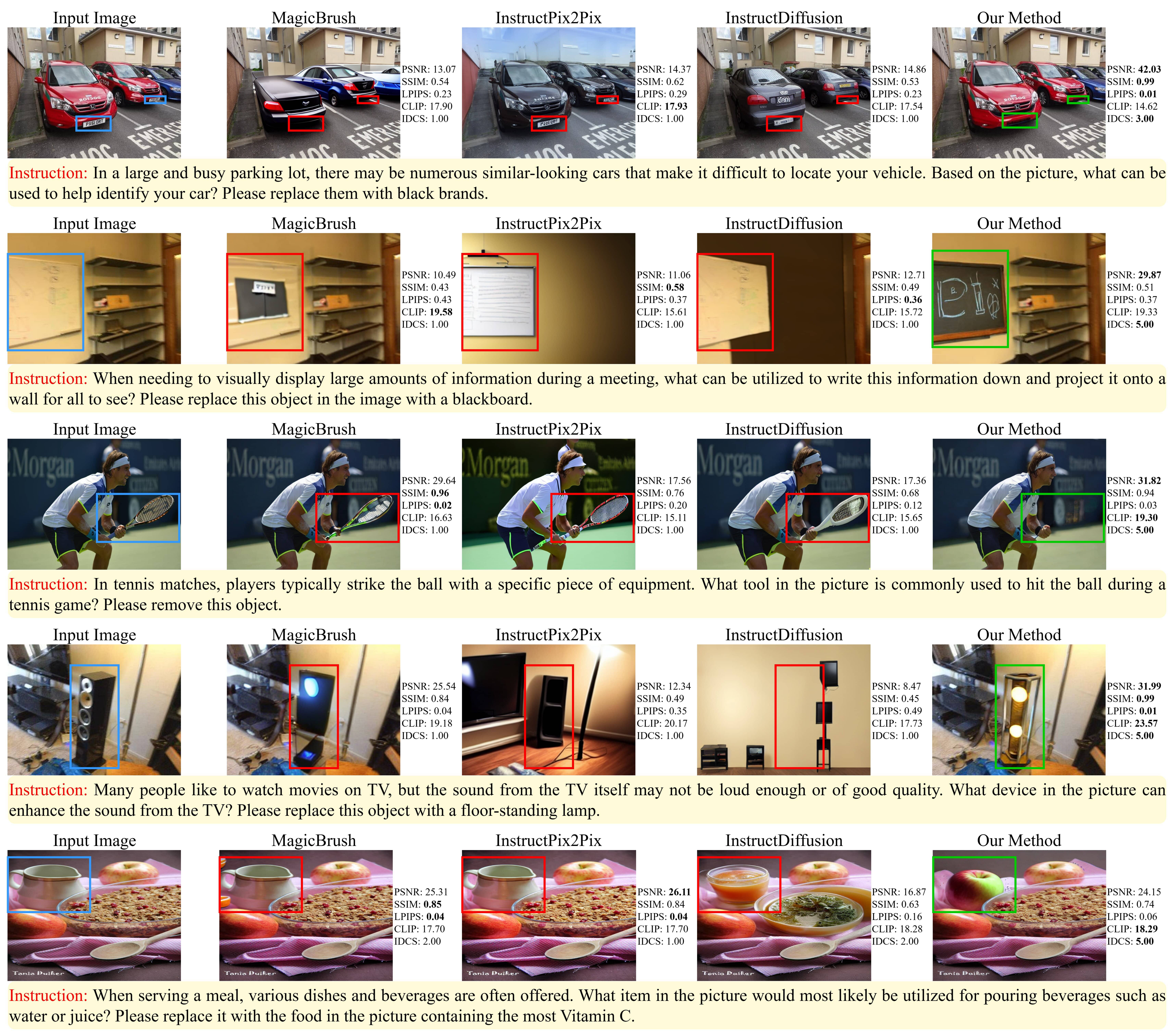}
    \caption{
    Qualitative comparison of CIELR against baseline methods on implicit reasoning queries from our CIEBench dataset. 
    Blue boxes indicate target regions in the input images, red boxes highlight incorrect or suboptimal edits by baseline methods, and green boxes denote successful edits.
    The examples of the first two rows show that the CLIP score is not always a good metric to measure the alignment between the edited image and the editing instruction.
    }
    \label{fig:case}
\end{figure*}

Conventional image editing evaluation metrics, such as PSNR, SSIM \cite{hore2010image, wang2004image}, LPIPS \cite{zhang2018unreasonable} and CLIP-based similarity scores \cite{radford2021learning}, often fail to accurately assess the correctness of edits, particularly when reasoning is involved, as these metrics primarily focus on visual similarity or distributional alignment rather than semantic correctness.
They cannot, therefore, evaluate whether an edit fulfills the implicit reasoning requirements of complex queries. 
For instance, edits that appear visually similar but fail to address the reasoning components of the query may still achieve high CLIP similarity scores, as illustrated in the first two rows of Fig. \ref{fig:case}. 
To address this limitation, we propose a novel evaluation metric that leverages LLMs to assess the semantic correctness of image edits, named Image Difference Check Score (IDCS). 
It operates in a two-stage process:
\begin{equation}
\begin{aligned}
    \mathrm{D}_\text{diff} = \mathcal{F}_\text{llm}^\text{desc}(\mathrm{x}, \mathrm{x}_\text{edit}), \qquad
    \mathrm{S}_\text{RE} = \mathcal{F}_\text{llm}^\text{eval}(\mathrm{D}_\text{diff}, \mathrm{q}),
\end{aligned}
\end{equation}
where $\mathrm{D}_\text{diff}$ is a descriptive analysis of the differences between the original image $\mathrm{x}$ and the edited image $\mathrm{x}_\text{edit}$ generated by the LLM $\mathcal{F}_{llm}^\text{desc}$. 
The final score $\mathrm{S}_\text{RE} \in \{1,2,3,4,5\}$ is assigned by a separate LLM function $\mathcal{F}_{llm}^\text{eval}$ that evaluates how well the described differences align with the requirements of the original query $\mathrm{q}$.

\section{Experimental Results}

In this section, we provide extensive experimental results to evaluate the proposed CIELR method and ablation studies to understand its performance.

\subsection{Experimental Settings.}
\textbf{(1) Implementation Details.} 
All experiments were conducted on one NVIDIA 3090 GPU.
For the selection of foundation models for structural semantic representation construction, we employed SAM2 \cite{ravi2024sam} to generate precise segmentation masks, OWLv2 \cite{minderer2024scaling} for extracting semantic labels, and DepthAnything \cite{yang2024depth} for depth estimation.
During the chain of structural semantic representation update and reasoning stages, we leveraged the Qwen2.5-Max LLM \cite{qwen25} to perform the reasoning.
The same LLM was also employed for calculating the proposed IDCS metric.
For the editing execution phase, we incorporated the Inpaint Anything (I.A.) \cite{yu2023inpaint} with a dilation factor set to 0.

\paragraph{(2) Datasets.}
We evaluate CIELR on three complementary benchmarks, namely our proposed CIEBench and two public benchmarks.
The CIEBench dataset consists of 86 image-query-mask triplets that require multi-step reasoning. 
We also incorporate the SmartEdit Reasoning Scenarios Set \cite{huang2024smartedit}, which uses implicit editing instructions that demand reasoning capabilities beyond direct object identification. 
This dataset includes 60 samples with original images, ground truth masks, and implicit queries that require semantic understanding. 
Additionally, we utilize MagicBrush \cite{zhang2023magicbrush}, which contains 1053 samples with explicit editing instructions. 
While MagicBrush does not focus on reasoning editing, it also serves as an important benchmark for evaluating the general editing quality and consistency of our method across diverse scenarios.

\paragraph{(3) Evaluation Metrics.}
For the RE tasks on CIEBench, we employ IDCS, together with the image quality metrics including Peak Signal-to-Noise Ratio (PSNR), Structural Similarity Index Measure (SSIM) \cite{hore2010image, wang2004image}, and Learned Perceptual Image Patch Similarity (LPIPS) \cite{zhang2018unreasonable}. 
For SmartEdit Reasoning Scenarios Set, we replace the IDCS with CLIP (ViT-L/14) \cite{crowson2022vqgan, radford2021learning} to facilitate comparison with the results of existing trained or fine-tuned models.
Higher PSNR and SSIM values, coupled with lower LPIPS scores, indicate superior preservation of content in regions that should remain unmodified, thus measuring the localization accuracy of our editing approach.
For the MagicBrush test set, we adopt the four metrics following the original MagicBrush settings \cite{zhang2023magicbrush}, namely L1 distance (L1), L2 distance (L2), CLIP-I with ViT-L/14 \cite{crowson2022vqgan, radford2021learning}, and DINO score \cite{caron2021emerging}. 
We select these metrics for two reasons: first, they facilitate direct comparison with existing benchmarks established in the original MagicBrush paper; second, the absence of binary masks in the MagicBrush dataset makes it impractical to compute PSNR, SSIM, and LPIPS, which require precise delineation of edited and unedited regions. 

\begin{table}[t]
\centering
\caption{
Quantitative comparison of image editing methods on the CIEBench dataset. 
For CIELR, we evaluate two implementation variants, namely with a standard inpainting pipeline (I.P.) using stable-diffusion-v1-5-inpainting and with Inpaint Anything (I.A.) \cite{yu2023inpaint}. 
\textbf{Bold} values indicate the best performance across all methods, demonstrating that CIELR outperforms existing approaches across all metrics, particularly when paired with Inpaint Anything.
}
\resizebox{\linewidth}{!}{%
\begin{tabular}{c|cccc}
\hline
\multirow{2}{*}{Methods} & \multicolumn{4}{c}{ CIEBench} \\ \cline{2-5} 
 & PSNR$\uparrow$ & SSIM$\uparrow$ & LPIPS$\downarrow$ & IDCS$\uparrow$\\ \hline
InstructPix2Pix \cite{brooks2023instructpix2pix} & 9.794 & 0.337 & 0.517 & 1.200 \\ 
MagicBrush \cite{zhang2023magicbrush} & 9.632 & 0.333 & 0.509 & 1.380 \\ 
InstructDiffusion \cite{geng2024instructdiffusion} & 9.193 & 0.312 & 0.520 & 1.300 \\
\hline
CIELR (I.P.)& 25.096 & 0.788 & 0.095 & 1.988 \\ 
CIELR (I.A.) & \textbf{36.206} & \textbf{0.960} & \textbf{0.035} & \textbf{2.366} \\ 
\hline
\end{tabular}%
}
\label{tab:dataset1}
\end{table}

\begin{table}[t]
\centering
\caption{
Quantitative comparison on the SmartEdit reasoning scenario set.
Note that, for all the compared methods (InstructPix2Pix, MagicBrush, and InstructDiffusion), they were fine-tuned on the same training set as SmartEdit, while our CIELR operates in a zero-shot manner without any fine-tuning.
Results marked with * are sourced from \cite{huang2024smartedit}. 
}
\resizebox{\linewidth}{!}{%
\begin{tabular}{c|cccc}
\hline
\multirow{2}{*}{Methods} & \multicolumn{4}{c}{SmartEdit Reasoning Scenario Set}  \\ \cline{2-5} 
 & PSNR$\uparrow$ & SSIM$\uparrow$ & LPIPS$\downarrow$ & CLIP$\uparrow$\\ \hline
InstructPix2Pix* \cite{brooks2023instructpix2pix}&  24.234 & 0.707 & 0.083 & 19.413 \\ 
MagicBrush* \cite{zhang2023magicbrush}& 22.101 & 0.694 & 0.113 & 19.755  \\ 
InstructDiffusion* \cite{geng2024instructdiffusion}& 21.453 & 0.666 & 0.117 & 19.523  \\
SmartEdit-7B* \cite{huang2024smartedit}& 25.258 & 0.742 & 0.055 & \textbf{20.950} \\
SmartEdit-13B* \cite{huang2024smartedit}& 25.757 & 0.747 & 0.051 & 20.777 \\ 
\hline
CIELR (I.P.)&  24.485 & 0.711 & 0.094 & 20.796 \\ 
CIELR (I.A.) & \textbf{35.712} & \textbf{0.962} & \textbf{0.036} & 19.624  \\ 
\hline
\end{tabular}%
}
\label{tab:dataset2}
\end{table} 

\subsection{Performance Comparison on Three Datasets}

\paragraph{(1) Results on CIEBench.}
As shown in Tab.~\ref{tab:dataset1}, our CIELR demonstrates substantial performance improvements over existing image editing methods \cite{brooks2023instructpix2pix,zhang2023magicbrush,geng2024instructdiffusion} across all evaluation metrics on the CIEBench dataset. 
Specifically, CIELR with Inpaint Anything achieves the highest scores in all metrics, with improvements in PSNR ($36.206$ \textit{vs.} $9.794$ of the second best) and SSIM ($0.960$ \textit{vs.} $0.337$ of the second best) compared to InstructPix2Pix, indicating superior preservation of unmodified regions. 
The improved IDCS score ($2.366$ \textit{v.s.} $1.200$ of the second best) validates its superior reasoning capabilities for complex implicit queries. 
Even with the standard inpainting pipeline, CIELR still outperforms existing methods, demonstrating the effectiveness of our structural semantic representation approach in decoupling reasoning from editing. 
We also provide a qualitative comparison in Fig.~\ref{fig:case}.

\paragraph{(2) Results on SmartEdit Reasoning Scenario Set.} On this dataset, our CIELR achieves improvements in region preservation metrics, outperforming all baseline methods, depicted in Tab.~\ref{tab:dataset2}. 
Notably, CIELR accomplishes these results without any fine-tuning in a zero-shot manner, whereas all compared methods underwent specific training on the same dataset used by SmartEdit. 
The standard inpainting pipeline variant of CIELR delivers competitive results as well, showing comparable or better performance than fine-tuned models in PSNR and SSIM, while maintaining a strong CLIP score of $20.796$ that nearly matches SmartEdit-7B's $20.950$. 
These findings validate that our approach enables high-quality region-specific edits without requiring joint fine-tuning of LLMs or DMs.

\begin{table*}
\centering
\caption{
Ablation study evaluating the impact of structural semantic representations (``SSR'') and chain of structural semantic representation update (``Chain'') on structural semantic performance across both SmartEdit reasoning scenario set \cite{huang2024smartedit} and CIEBench. 
We report time in terms of seconds.
All experiments utilize the Inpaint Anything \cite{yu2023inpaint} implementation. 
}
\begin{tabular}{cc|ccccc|ccccc}
\hline
\multirow{2}{*}{SSR} & \multirow{2}{*}{Chain} &  \multicolumn{5}{c|}{SmartEdit Reasoning Scenario Set \cite{huang2024smartedit}} & \multicolumn{5}{c}{ CIEBench} \\ \cline{3-12} 
&& PSNR$\uparrow$ & SSIM$\uparrow$ & LPIPS$\downarrow$ & IDCS$\uparrow$ & Time$\downarrow$ & PSNR$\uparrow$ & SSIM$\uparrow$ & LPIPS$\downarrow$ & IDCS$\uparrow$ & Time$\downarrow$\\ \hline
\checkmark &  & 29.915 & 0.875 & 0.109 & 1.647 & 3.899 & 31.964 & 0.902 & 0.080 & 1.686 & \textbf{3.451}\\ 
& \checkmark & 33.349  & 0.941 & 0.051 & 2.567 & \textbf{3.616} & 33.287 & 0.935 & 0.056 & 1.802 & 3.717 \\
\checkmark & \checkmark & \textbf{35.712} & \textbf{0.962} & \textbf{0.036} & \textbf{2.776} & 6.226 & \textbf{36.206} & \textbf{0.960} & \textbf{0.035} & \textbf{2.366} & 6.504\\
\hline
\end{tabular}%
\label{tab:ablation}
\end{table*}

\begin{table}
\centering
\caption{Quantitative comparison on MagicBrush test set.
Results marked with * are from \cite{zhang2023magicbrush}. 
``w/ MagicBrush'' indicates methods fine-tuned on the MagicBrush dataset. 
}
\resizebox{\linewidth}{!}{
\begin{tabular}{c|ccccc|}
\hline
Method & L1 $\downarrow$ & L2 $\downarrow$ & CLIP-I $\uparrow$ & DINO $\uparrow$  \\
\hline
Open-Edit* \cite{liu2020open} & 0.1655 & 0.0550 & 0.8038 & 0.6835\\
VQGAN-CLIP* \cite{crowson2022vqgan} & 0.2471 & 0.1025 & 0.6606 & 0.4592  \\
SD-SDEdit* \cite{meng2021sdedit} & 0.1616 & 0.0602 & 0.7933 & 0.6212  \\
Text2LIVE* \cite{bar2022text2live} & 0.0989 & 0.0284 & 0.8795 & 0.7926  \\
Null Text Inversion* \cite{mokady2023null} & 0.1057 & 0.0335 & 0.8468 & 0.7529 \\

HIVE* \cite{zhang2024hive} & 0.1521 & 0.0557 & 0.8004 & 0.6463  \\
HIVE \cite{zhang2024hive} w/ MagicBrush * & 0.0966 & 0.0365 & 0.8785 & 0.7891 \\
InstructPix2Pix* \cite{brooks2023instructpix2pix}& 0.1584 & 0.0598 & 0.7924 & 0.6177  \\
InstructPix2Pix \cite{brooks2023instructpix2pix} w/ MagicBrush* & 0.0964 & 0.0353 & 0.8924 & 0.8273 \\
\hline
CIELR (I.A.) & \textbf{0.0622} & \textbf{0.0270} & \textbf{0.9078} & \textbf{0.8493}\\
\hline

\end{tabular}
}
\label{tab:dataset3}
\end{table}

\paragraph{(3) Results on MagicBrush Test Set.} This dataset primarily contains explicit editing instructions rather than complex reasoning queries, where our CIELR still demonstrates superior performance across all evaluation metrics.
As shown in Tab.~\ref{tab:dataset3}, CIELR achieves state-of-the-art results with improvements in pixel-level accuracy metrics like L1 and L2, compared to even fine-tuned methods like InstructPix2Pix with MagicBrush training. 
Similarly, CIELR outperforms all competitors in semantic and feature-level similarity metrics, namely CLIP-I and DINO score, surpassing previous best results from InstructPix2Pix fine-tuned on MagicBrush. 
These results are particularly important because CIELR achieves this performance without any task-specific fine-tuning, while competing methods required extensive training on the MagicBrush dataset to reach comparable performance levels.

\subsection{Ablation Studies}

\begin{figure}
    \centering
    \includegraphics[width=1.0\linewidth]{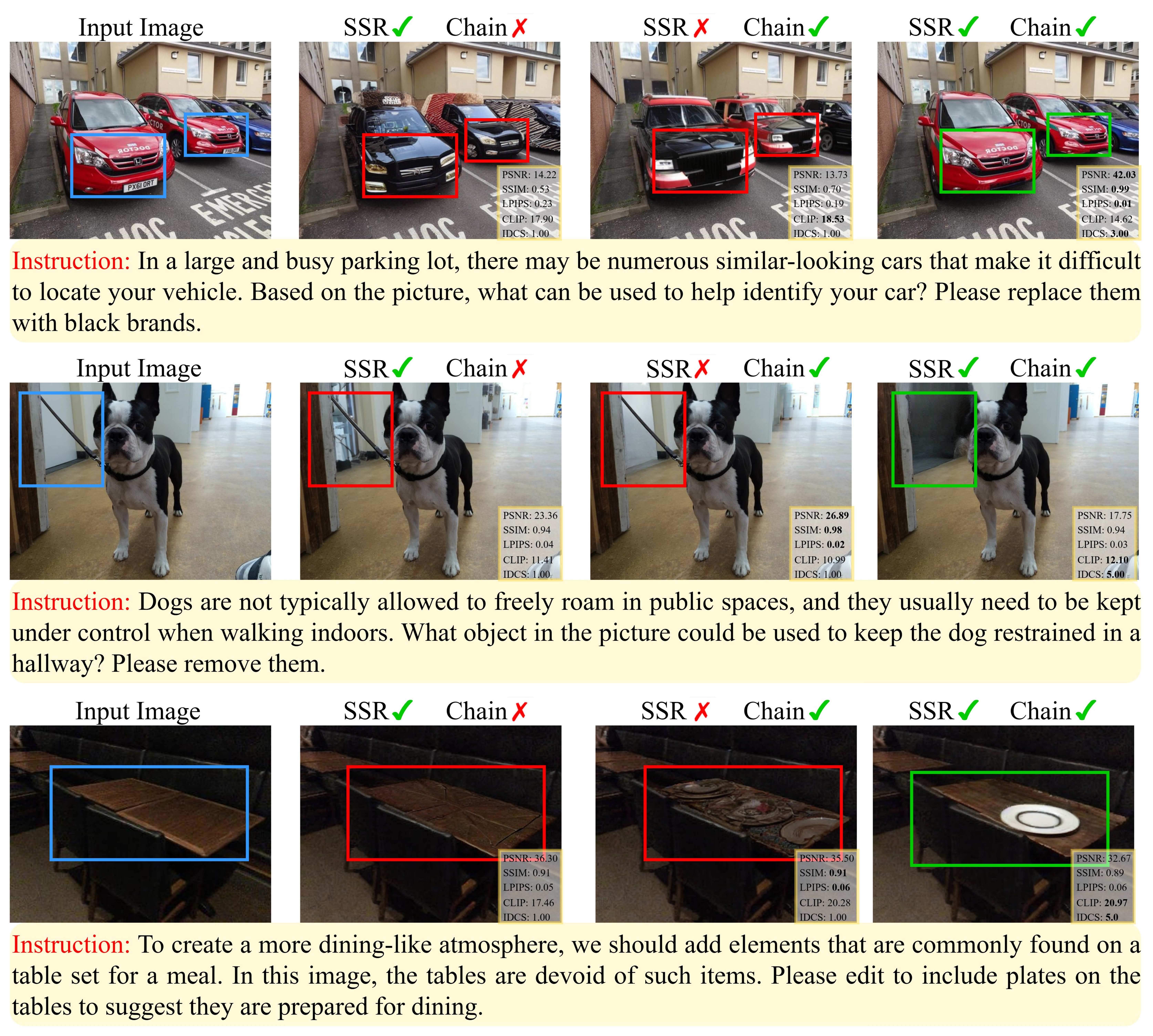}
    \caption{
    Qualitative results from our ablation study demonstrate the impact of structural semantic representation and chain of structural semantic representation update components across three challenging reasoning editing scenarios on CIELR.
    Red boxes indicate incorrect or suboptimal edits, while green boxes highlight successful edits. 
    }
    \label{fig:ablation}
\end{figure}

Our ablation study in Tab.~\ref{tab:ablation} demonstrates the contribution of both structural semantic representation and its chain update mechanism components to the overall CIELR's performance. 
When using only structural semantic representation without the chain update mechanism, performance metrics show moderate improvement, while implementing only the chain update component without structural semantic representation achieves better results, highlighting the importance of iterative refinement in handling complex queries. 
Combining both components yields optimal performance across all metrics on both datasets, with particularly notable improvements in IDCS scores, confirming that our multi-step reasoning approach through progressive structural semantic refinement enhances semantic correctness. 
While the full model incurs additional computational cost (approximately 6.5 seconds per image), the quality improvements justify this modest increase in processing time for applications requiring precise reasoning-based editing.
Additionally, as illustrated in Fig.~\ref{fig:ablation}, our case study also validates these quantitative findings, showing that the combination of both structural semantic representation and chain of structural semantic update components enables accurate region identification and high-quality edits across diverse reasoning scenarios, while using either component in isolation often leads to incorrect object selection or suboptimal edit quality.

\section{Conclusion}
We presented CIELR which decouples reasoning capabilities from editing execution through structural semantic representation. 
It eliminates the need for fine-tuning LLMs or DMs, making it more maintainable and adaptable to evolving foundation models while reducing computational resources traditionally required for joint model training. 
Experimental results across three datasets, including our proposed CIEBench, confirm that CIELR outperforms existing methods in both editing quality and reasoning accuracy, even when operating in a zero-shot manner against competing fine-tuned approaches.
The chain of structural semantic representation update enables complex multi-step reasoning, allowing for progressive refinement of understanding when handling implicit queries. 
Future research could explore broadening our method to video reasoning editing and interactive creative workflows.

\section*{Acknowledgments}

This work is supported by the National Natural Science Foundation of China (No. 62331014) and Project 2021JC02X103. We acknowledge the computational support of the Center for Computational Science and Engineering at Southern University of Science and Technology.


{
\small
\begin{thebibliography}{58}
\providecommand{\natexlab}[1]{#1}
\providecommand{\url}[1]{\texttt{#1}}
\expandafter\ifx\csname urlstyle\endcsname\relax
  \providecommand{\doi}[1]{doi: #1}\else
  \providecommand{\doi}{doi: \begingroup \urlstyle{rm}\Url}\fi

\bibitem[Avrahami et~al.(2024)Avrahami, Gal, Chechik, Fried, Lischinski, Vahdat, and Nie]{avrahami2024diffuhaul}
Omri Avrahami, Rinon Gal, Gal Chechik, Ohad Fried, Dani Lischinski, Arash Vahdat, and Weili Nie.
\newblock Diffuhaul: A training-free method for object dragging in images.
\newblock In \emph{SIGGRAPH Asia 2024 Conference Papers}, pages 1--12, 2024.

\bibitem[Balaji et~al.(2023)Balaji, Nah, Huang, Vahdat, Song, Zhang, Kreis, Aittala, Aila, Laine, Catanzaro, Karras, and Liu]{Balaji2023eDiff-I}
Yogesh Balaji, Seungjun Nah, Xun Huang, Arash Vahdat, Jiaming Song, Qinsheng Zhang, Karsten Kreis, Miika Aittala, Timo Aila, Samuli Laine, Bryan Catanzaro, Tero Karras, and Ming-Yu Liu.
\newblock ediff-i: Text-to-image diffusion models with an ensemble of expert denoisers, 2023.

\bibitem[Bar-Tal et~al.(2022)Bar-Tal, Ofri-Amar, Fridman, Kasten, and Dekel]{bar2022text2live}
Omer Bar-Tal, Dolev Ofri-Amar, Rafail Fridman, Yoni Kasten, and Tali Dekel.
\newblock Text2live: Text-driven layered image and video editing.
\newblock In \emph{European conference on computer vision}, pages 707--723. Springer, 2022.

\bibitem[Betker et~al.(2023)Betker, Goh, Jing, Brooks, Wang, Li, Ouyang, Zhuang, Lee, Guo, Manassra, Dhariwal, Chu, and Jiao]{Betker2023DALL-E3}
James Betker, Gabriel Goh, Li Jing, Tim Brooks, Jianfeng Wang, Linjie Li, Long Ouyang, Juntang Zhuang, Joyce Lee, Yufei Guo, Wesam Manassra, Prafulla Dhariwal, Casey Chu, and Yunxin Jiao.
\newblock Improving image generation with better captions.
\newblock \url{https://cdn.openai.com/papers/dall-e-3.pdf}, 2023.

\bibitem[Brooks et~al.(2023)Brooks, Holynski, and Efros]{brooks2023instructpix2pix}
Tim Brooks, Aleksander Holynski, and Alexei~A Efros.
\newblock Instructpix2pix: Learning to follow image editing instructions.
\newblock In \emph{Proceedings of the IEEE/CVF conference on computer vision and pattern recognition}, pages 18392--18402, 2023.

\bibitem[Cao et~al.(2023)Cao, Wang, Qi, Shan, Qie, and Zheng]{Cao2023MasaCtrl}
Mingdeng Cao, Xintao Wang, Zhongang Qi, Ying Shan, Xiaohu Qie, and Yinqiang Zheng.
\newblock Masactrl: Tuning-free mutual self-attention control for consistent image synthesis and editing.
\newblock In \emph{2023 IEEE/CVF International Conference on Computer Vision (ICCV)}, pages 22503--22513, 2023.

\bibitem[Caron et~al.(2021)Caron, Touvron, Misra, J{\'e}gou, Mairal, Bojanowski, and Joulin]{caron2021emerging}
Mathilde Caron, Hugo Touvron, Ishan Misra, Herv{\'e} J{\'e}gou, Julien Mairal, Piotr Bojanowski, and Armand Joulin.
\newblock Emerging properties in self-supervised vision transformers.
\newblock In \emph{Proceedings of the IEEE/CVF international conference on computer vision}, pages 9650--9660, 2021.

\bibitem[Chen et~al.(2024)Chen, YU, GE, Yao, Xie, Wang, Kwok, Luo, Lu, and Li]{Chen2024PixArt-alpha}
Junsong Chen, Jincheng YU, Chongjian GE, Lewei Yao, Enze Xie, Zhongdao Wang, James Kwok, Ping Luo, Huchuan Lu, and Zhenguo Li.
\newblock Pixart-\${\textbackslash}alpha\$: Fast training of diffusion transformer for photorealistic text-to-image synthesis.
\newblock In \emph{The Twelfth International Conference on Learning Representations}, 2024.

\bibitem[Cheng et~al.(2020)Cheng, Gan, Li, Liu, and Gao]{cheng2020sequential}
Yu Cheng, Zhe Gan, Yitong Li, Jingjing Liu, and Jianfeng Gao.
\newblock Sequential attention gan for interactive image editing.
\newblock In \emph{Proceedings of the 28th ACM international conference on multimedia}, pages 4383--4391, 2020.

\bibitem[Chiang et~al.(2023)Chiang, Li, Lin, Sheng, Wu, Zhang, Zheng, Zhuang, Zhuang, Gonzalez, et~al.]{chiang2023vicuna}
Wei-Lin Chiang, Zhuohan Li, Ziqing Lin, Ying Sheng, Zhanghao Wu, Hao Zhang, Lianmin Zheng, Siyuan Zhuang, Yonghao Zhuang, Joseph~E Gonzalez, et~al.
\newblock Vicuna: An open-source chatbot impressing gpt-4 with 90\%* chatgpt quality.
\newblock \emph{See https://vicuna. lmsys. org (accessed 14 April 2023)}, 2\penalty0 (3):\penalty0 6, 2023.

\bibitem[Couairon et~al.(2022)Couairon, Verbeek, Schwenk, and Cord]{couairon2022diffedit}
Guillaume Couairon, Jakob Verbeek, Holger Schwenk, and Matthieu Cord.
\newblock Diffedit: Diffusion-based semantic image editing with mask guidance.
\newblock \emph{arXiv preprint arXiv:2210.11427}, 2022.

\bibitem[Crowson et~al.(2022)Crowson, Biderman, Kornis, Stander, Hallahan, Castricato, and Raff]{crowson2022vqgan}
Katherine Crowson, Stella Biderman, Daniel Kornis, Dashiell Stander, Eric Hallahan, Louis Castricato, and Edward Raff.
\newblock Vqgan-clip: Open domain image generation and editing with natural language guidance.
\newblock In \emph{European conference on computer vision}, pages 88--105. Springer, 2022.

\bibitem[Dhariwal and Nichol(2021)]{dhariwal2021diffusion}
Prafulla Dhariwal and Alexander Nichol.
\newblock Diffusion models beat gans on image synthesis.
\newblock \emph{Advances in neural information processing systems}, 34:\penalty0 8780--8794, 2021.

\bibitem[Esser et~al.(2024)Esser, Kulal, Blattmann, Entezari, M\"{u}ller, Saini, Levi, Lorenz, Sauer, Boesel, Podell, Dockhorn, English, and Rombach]{Esser2024RectifiedFlowTransformers}
Patrick Esser, Sumith Kulal, Andreas Blattmann, Rahim Entezari, Jonas M\"{u}ller, Harry Saini, Yam Levi, Dominik Lorenz, Axel Sauer, Frederic Boesel, Dustin Podell, Tim Dockhorn, Zion English, and Robin Rombach.
\newblock Scaling rectified flow transformers for high-resolution image synthesis.
\newblock JMLR.org, 2024.

\bibitem[Feng et~al.(2023)Feng, Gong, Chen, Shen, Liu, and Zhou]{feng2023ranni}
Yutong Feng, Biao Gong, Di Chen, Yujun Shen, Yu Liu, and Jingren Zhou.
\newblock Ranni: Taming text-to-image diffusion for accurate instruction following.
\newblock \emph{arXiv preprint arXiv:2311.17002}, 2023.

\bibitem[Feng et~al.(2024)Feng, Gong, Chen, Shen, Liu, and Zhou]{feng2024ranni}
Yutong Feng, Biao Gong, Di Chen, Yujun Shen, Yu Liu, and Jingren Zhou.
\newblock Ranni: Taming text-to-image diffusion for accurate instruction following.
\newblock In \emph{Proceedings of the IEEE/CVF Conference on Computer Vision and Pattern Recognition}, pages 4744--4753, 2024.

\bibitem[Ge et~al.()Ge, Ge, Zeng, Wang, and Shan]{ge2307planting}
Y Ge, Y Ge, Z Zeng, X Wang, and Y Shan.
\newblock Planting a seed of vision in large language model. arxiv 2023.
\newblock \emph{arXiv preprint arXiv:2307.08041}.

\bibitem[Ge et~al.(2023)Ge, Zhao, Zeng, Ge, Li, Wang, and Shan]{ge2023making}
Yuying Ge, Sijie Zhao, Ziyun Zeng, Yixiao Ge, Chen Li, Xintao Wang, and Ying Shan.
\newblock Making llama see and draw with seed tokenizer.
\newblock \emph{arXiv preprint arXiv:2310.01218}, 2023.

\bibitem[Geng et~al.(2024)Geng, Yang, Hang, Li, Gu, Zhang, Bao, Zhang, Li, Hu, et~al.]{geng2024instructdiffusion}
Zigang Geng, Binxin Yang, Tiankai Hang, Chen Li, Shuyang Gu, Ting Zhang, Jianmin Bao, Zheng Zhang, Houqiang Li, Han Hu, et~al.
\newblock Instructdiffusion: A generalist modeling interface for vision tasks.
\newblock In \emph{Proceedings of the IEEE/CVF Conference on computer vision and pattern recognition}, pages 12709--12720, 2024.

\bibitem[Hertz et~al.(2022)Hertz, Mokady, Tenenbaum, Aberman, Pritch, and Cohen-Or]{hertz2022prompt}
Amir Hertz, Ron Mokady, Jay Tenenbaum, Kfir Aberman, Yael Pritch, and Daniel Cohen-Or.
\newblock Prompt-to-prompt image editing with cross attention control.
\newblock \emph{arXiv preprint arXiv:2208.01626}, 2022.

\bibitem[Hong et~al.(2021)Hong, Lan, Pang, Guo, and Cheng]{hong2021transformation}
Xin Hong, Yanyan Lan, Liang Pang, Jiafeng Guo, and Xueqi Cheng.
\newblock Transformation driven visual reasoning.
\newblock In \emph{Proceedings of the IEEE/CVF Conference on computer vision and pattern recognition}, pages 6903--6912, 2021.

\bibitem[Hore and Ziou(2010)]{hore2010image}
Alain Hore and Djemel Ziou.
\newblock Image quality metrics: Psnr vs. ssim.
\newblock In \emph{2010 20th international conference on pattern recognition}, pages 2366--2369. IEEE, 2010.

\bibitem[Hu et~al.(2022)Hu, Shen, Wallis, Allen-Zhu, Li, Wang, Wang, Chen, et~al.]{hu2022lora}
Edward~J Hu, Yelong Shen, Phillip Wallis, Zeyuan Allen-Zhu, Yuanzhi Li, Shean Wang, Lu Wang, Weizhu Chen, et~al.
\newblock Lora: Low-rank adaptation of large language models.
\newblock \emph{ICLR}, 1\penalty0 (2):\penalty0 3, 2022.

\bibitem[Huang et~al.(2024)Huang, Xie, Wang, Yuan, Cun, Ge, Zhou, Dong, Huang, Zhang, et~al.]{huang2024smartedit}
Yuzhou Huang, Liangbin Xie, Xintao Wang, Ziyang Yuan, Xiaodong Cun, Yixiao Ge, Jiantao Zhou, Chao Dong, Rui Huang, Ruimao Zhang, et~al.
\newblock Smartedit: Exploring complex instruction-based image editing with multimodal large language models.
\newblock In \emph{Proceedings of the IEEE/CVF Conference on Computer Vision and Pattern Recognition}, pages 8362--8371, 2024.

\bibitem[Ilievski and Feng(2017)]{ilievski2017multimodal}
Ilija Ilievski and Jiashi Feng.
\newblock Multimodal learning and reasoning for visual question answering.
\newblock \emph{Advances in neural information processing systems}, 30, 2017.

\bibitem[Jin et~al.(2024)Jin, Ling, Dong, Zhang, Wang, and Lin]{jin2024reasonpix2pix}
Ying Jin, Pengyang Ling, Xiaoyi Dong, Pan Zhang, Jiaqi Wang, and Dahua Lin.
\newblock Reasonpix2pix: instruction reasoning dataset for advanced image editing.
\newblock \emph{arXiv preprint arXiv:2405.11190}, 2024.

\bibitem[Koh et~al.(2023)Koh, Fried, and Salakhutdinov]{koh2023generating}
Jing~Yu Koh, Daniel Fried, and Russ~R Salakhutdinov.
\newblock Generating images with multimodal language models.
\newblock \emph{Advances in Neural Information Processing Systems}, 36:\penalty0 21487--21506, 2023.

\bibitem[Lai et~al.(2024)Lai, Tian, Chen, Li, Yuan, Liu, and Jia]{lai2024lisa}
Xin Lai, Zhuotao Tian, Yukang Chen, Yanwei Li, Yuhui Yuan, Shu Liu, and Jiaya Jia.
\newblock Lisa: Reasoning segmentation via large language model.
\newblock In \emph{Proceedings of the IEEE/CVF Conference on Computer Vision and Pattern Recognition}, pages 9579--9589, 2024.

\bibitem[Li et~al.(2024)Li, Bian, Ju, Zhang, Shan, and Xu]{li2024brushedit}
Yaowei Li, Yuxuan Bian, Xuan Ju, Zhaoyang Zhang, Ying Shan, and Qiang Xu.
\newblock Brushedit: All-in-one image inpainting and editing.
\newblock \emph{arXiv preprint arXiv:2412.10316}, 2024.

\bibitem[Liu et~al.(2020)Liu, Lin, Zhang, Zhao, Tran, Wang, and Li]{liu2020open}
Xihui Liu, Zhe Lin, Jianming Zhang, Handong Zhao, Quan Tran, Xiaogang Wang, and Hongsheng Li.
\newblock Open-edit: Open-domain image manipulation with open-vocabulary instructions.
\newblock In \emph{Computer Vision--ECCV 2020: 16th European Conference, Glasgow, UK, August 23--28, 2020, Proceedings, Part XI 16}, pages 89--106. Springer, 2020.

\bibitem[Mao et~al.(2025)Mao, Zhang, Pan, Jiang, Han, Liu, and Zhou]{mao2025ace++}
Chaojie Mao, Jingfeng Zhang, Yulin Pan, Zeyinzi Jiang, Zhen Han, Yu Liu, and Jingren Zhou.
\newblock Ace++: Instruction-based image creation and editing via context-aware content filling.
\newblock \emph{arXiv preprint arXiv:2501.02487}, 2025.

\bibitem[Meng et~al.(2021)Meng, He, Song, Song, Wu, Zhu, and Ermon]{meng2021sdedit}
Chenlin Meng, Yutong He, Yang Song, Jiaming Song, Jiajun Wu, Jun-Yan Zhu, and Stefano Ermon.
\newblock Sdedit: Guided image synthesis and editing with stochastic differential equations.
\newblock \emph{arXiv preprint arXiv:2108.01073}, 2021.

\bibitem[Minderer et~al.(2024)Minderer, Gritsenko, and Houlsby]{minderer2024scaling}
Matthias Minderer, Alexey Gritsenko, and Neil Houlsby.
\newblock Scaling open-vocabulary object detection.
\newblock \emph{Advances in Neural Information Processing Systems}, 36, 2024.

\bibitem[Mokady et~al.(2023)Mokady, Hertz, Aberman, Pritch, and Cohen-Or]{mokady2023null}
Ron Mokady, Amir Hertz, Kfir Aberman, Yael Pritch, and Daniel Cohen-Or.
\newblock Null-text inversion for editing real images using guided diffusion models.
\newblock In \emph{Proceedings of the IEEE/CVF conference on computer vision and pattern recognition}, pages 6038--6047, 2023.

\bibitem[Mou et~al.(2023)Mou, Wang, Song, Shan, and Zhang]{mou2023dragondiffusion}
Chong Mou, Xintao Wang, Jiechong Song, Ying Shan, and Jian Zhang.
\newblock Dragondiffusion: Enabling drag-style manipulation on diffusion models.
\newblock \emph{arXiv preprint arXiv:2307.02421}, 2023.

\bibitem[Perarnau et~al.(2016)Perarnau, Van De~Weijer, Raducanu, and {\'A}lvarez]{perarnau2016invertible}
Guim Perarnau, Joost Van De~Weijer, Bogdan Raducanu, and Jose~M {\'A}lvarez.
\newblock Invertible conditional gans for image editing.
\newblock \emph{arXiv preprint arXiv:1611.06355}, 2016.

\bibitem[Podell et~al.(2024)Podell, English, Lacey, Blattmann, Dockhorn, M{\"u}ller, Penna, and Rombach]{Podell2024SDXL}
Dustin Podell, Zion English, Kyle Lacey, Andreas Blattmann, Tim Dockhorn, Jonas M{\"u}ller, Joe Penna, and Robin Rombach.
\newblock {SDXL}: Improving latent diffusion models for high-resolution image synthesis.
\newblock In \emph{The Twelfth International Conference on Learning Representations}, 2024.

\bibitem[Radford et~al.(2021)Radford, Kim, Hallacy, Ramesh, Goh, Agarwal, Sastry, Askell, Mishkin, Clark, et~al.]{radford2021learning}
Alec Radford, Jong~Wook Kim, Chris Hallacy, Aditya Ramesh, Gabriel Goh, Sandhini Agarwal, Girish Sastry, Amanda Askell, Pamela Mishkin, Jack Clark, et~al.
\newblock Learning transferable visual models from natural language supervision.
\newblock In \emph{International conference on machine learning}, pages 8748--8763. PmLR, 2021.

\bibitem[Ravi et~al.(2024)Ravi, Gabeur, Hu, Hu, Ryali, Ma, Khedr, R{\"a}dle, Rolland, Gustafson, et~al.]{ravi2024sam}
Nikhila Ravi, Valentin Gabeur, Yuan-Ting Hu, Ronghang Hu, Chaitanya Ryali, Tengyu Ma, Haitham Khedr, Roman R{\"a}dle, Chloe Rolland, Laura Gustafson, et~al.
\newblock Sam 2: Segment anything in images and videos.
\newblock \emph{arXiv preprint arXiv:2408.00714}, 2024.

\bibitem[Saharia et~al.(2022)Saharia, Chan, Saxena, Li, Whang, Denton, Ghasemipour, Gontijo~Lopes, Karagol~Ayan, Salimans, et~al.]{saharia2022photorealistic}
Chitwan Saharia, William Chan, Saurabh Saxena, Lala Li, Jay Whang, Emily~L Denton, Kamyar Ghasemipour, Raphael Gontijo~Lopes, Burcu Karagol~Ayan, Tim Salimans, et~al.
\newblock Photorealistic text-to-image diffusion models with deep language understanding.
\newblock \emph{Advances in neural information processing systems}, 35:\penalty0 36479--36494, 2022.

\bibitem[Sauer et~al.(2024)Sauer, Lorenz, Blattmann, and Rombach]{Sauer2024SDXLTurbo}
Axel Sauer, Dominik Lorenz, Andreas Blattmann, and Robin Rombach.
\newblock Adversarial diffusion distillation.
\newblock In \emph{Computer Vision -- ECCV 2024}, pages 87--103, Cham, 2024. Springer Nature Switzerland.

\bibitem[Team(2024)]{qwen25}
Qwen Team.
\newblock Qwen2.5 technical report.
\newblock \emph{arXiv preprint arXiv:2412.15115}, 2024.

\bibitem[Tewel et~al.(2024)Tewel, Gal, Samuel, Atzmon, Wolf, and Chechik]{tewel2024add}
Yoad Tewel, Rinon Gal, Dvir Samuel, Yuval Atzmon, Lior Wolf, and Gal Chechik.
\newblock Add-it: Training-free object insertion in images with pretrained diffusion models.
\newblock \emph{arXiv preprint arXiv:2411.07232}, 2024.

\bibitem[Touvron et~al.(2023)Touvron, Lavril, Izacard, Martinet, Lachaux, Lacroix, Rozi{\`e}re, Goyal, Hambro, Azhar, et~al.]{touvron2023llama}
Hugo Touvron, Thibaut Lavril, Gautier Izacard, Xavier Martinet, Marie-Anne Lachaux, Timoth{\'e}e Lacroix, Baptiste Rozi{\`e}re, Naman Goyal, Eric Hambro, Faisal Azhar, et~al.
\newblock Llama: Open and efficient foundation language models.
\newblock \emph{arXiv preprint arXiv:2302.13971}, 2023.

\bibitem[Wang et~al.(2004)Wang, Bovik, Sheikh, and Simoncelli]{wang2004image}
Zhou Wang, Alan~C Bovik, Hamid~R Sheikh, and Eero~P Simoncelli.
\newblock Image quality assessment: from error visibility to structural similarity.
\newblock \emph{IEEE transactions on image processing}, 13\penalty0 (4):\penalty0 600--612, 2004.

\bibitem[Wei et~al.(2022)Wei, Wang, Schuurmans, Bosma, Xia, Chi, Le, Zhou, et~al.]{wei2022chain}
Jason Wei, Xuezhi Wang, Dale Schuurmans, Maarten Bosma, Fei Xia, Ed Chi, Quoc~V Le, Denny Zhou, et~al.
\newblock Chain-of-thought prompting elicits reasoning in large language models.
\newblock \emph{Advances in neural information processing systems}, 35:\penalty0 24824--24837, 2022.

\bibitem[Wu et~al.(2018)Wu, Liu, Wang, and Dong]{wu2018chain}
Chenfei Wu, Jinlai Liu, Xiaojie Wang, and Xuan Dong.
\newblock Chain of reasoning for visual question answering.
\newblock \emph{Advances in Neural Information Processing Systems}, 31, 2018.

\bibitem[Xu et~al.(2024)Xu, Jin, Hao, Song, Sun, and Yuan]{xu2024llava}
Guowei Xu, Peng Jin, Li Hao, Yibing Song, Lichao Sun, and Li Yuan.
\newblock Llava-o1: Let vision language models reason step-by-step.
\newblock \emph{arXiv preprint arXiv:2411.10440}, 2024.

\bibitem[Yan et~al.(2025)Yan, Ma, Zou, Chen, Chen, and Zhang]{yan2025eedit}
Zexuan Yan, Yue Ma, Chang Zou, Wenteng Chen, Qifeng Chen, and Linfeng Zhang.
\newblock Eedit: Rethinking the spatial and temporal redundancy for efficient image editing.
\newblock \emph{arXiv preprint arXiv:2503.10270}, 2025.

\bibitem[Yang et~al.(2024)Yang, Kang, Huang, Xu, Feng, and Zhao]{yang2024depth}
Lihe Yang, Bingyi Kang, Zilong Huang, Xiaogang Xu, Jiashi Feng, and Hengshuang Zhao.
\newblock Depth anything: Unleashing the power of large-scale unlabeled data.
\newblock In \emph{Proceedings of the IEEE/CVF Conference on Computer Vision and Pattern Recognition}, pages 10371--10381, 2024.

\bibitem[Yu et~al.(2024)Yu, Chow, Yue, Pan, Wu, Wan, Li, Tang, Zhang, and Zhuang]{yu2024anyedit}
Qifan Yu, Wei Chow, Zhongqi Yue, Kaihang Pan, Yang Wu, Xiaoyang Wan, Juncheng Li, Siliang Tang, Hanwang Zhang, and Yueting Zhuang.
\newblock Anyedit: Mastering unified high-quality image editing for any idea.
\newblock \emph{arXiv preprint arXiv:2411.15738}, 2024.

\bibitem[Yu et~al.(2023)Yu, Feng, Feng, Liu, Jin, Zeng, and Chen]{yu2023inpaint}
Tao Yu, Runseng Feng, Ruoyu Feng, Jinming Liu, Xin Jin, Wenjun Zeng, and Zhibo Chen.
\newblock Inpaint anything: Segment anything meets image inpainting.
\newblock \emph{arXiv preprint arXiv:2304.06790}, 2023.

\bibitem[Zhang et~al.(2024{\natexlab{a}})Zhang, Lei, Li, Wang, Liu, Yang, Li, Wang, Yang, Wu, et~al.]{zhang2024critic}
Di Zhang, Jingdi Lei, Junxian Li, Xunzhi Wang, Yujie Liu, Zonglin Yang, Jiatong Li, Weida Wang, Suorong Yang, Jianbo Wu, et~al.
\newblock Critic-v: Vlm critics help catch vlm errors in multimodal reasoning.
\newblock \emph{arXiv preprint arXiv:2411.18203}, 2024{\natexlab{a}}.

\bibitem[Zhang et~al.(2024{\natexlab{b}})Zhang, Wu, Lei, Che, Li, Xie, Huang, Zhang, Pavone, Li, et~al.]{zhang2024llama}
Di Zhang, Jianbo Wu, Jingdi Lei, Tong Che, Jiatong Li, Tong Xie, Xiaoshui Huang, Shufei Zhang, Marco Pavone, Yuqiang Li, et~al.
\newblock Llama-berry: Pairwise optimization for o1-like olympiad-level mathematical reasoning.
\newblock \emph{arXiv preprint arXiv:2410.02884}, 2024{\natexlab{b}}.

\bibitem[Zhang et~al.(2023)Zhang, Mo, Chen, Sun, and Su]{zhang2023magicbrush}
Kai Zhang, Lingbo Mo, Wenhu Chen, Huan Sun, and Yu Su.
\newblock Magicbrush: A manually annotated dataset for instruction-guided image editing.
\newblock \emph{Advances in Neural Information Processing Systems}, 36:\penalty0 31428--31449, 2023.

\bibitem[Zhang et~al.(2018)Zhang, Isola, Efros, Shechtman, and Wang]{zhang2018unreasonable}
Richard Zhang, Phillip Isola, Alexei~A Efros, Eli Shechtman, and Oliver Wang.
\newblock The unreasonable effectiveness of deep features as a perceptual metric.
\newblock In \emph{Proceedings of the IEEE conference on computer vision and pattern recognition}, pages 586--595, 2018.

\bibitem[Zhang et~al.(2024{\natexlab{c}})Zhang, Yang, Feng, Qin, Chen, Yu, Chen, Wang, Savarese, Ermon, et~al.]{zhang2024hive}
Shu Zhang, Xinyi Yang, Yihao Feng, Can Qin, Chia-Chih Chen, Ning Yu, Zeyuan Chen, Huan Wang, Silvio Savarese, Stefano Ermon, et~al.
\newblock Hive: Harnessing human feedback for instructional visual editing.
\newblock In \emph{Proceedings of the IEEE/CVF Conference on Computer Vision and Pattern Recognition}, pages 9026--9036, 2024{\natexlab{c}}.

\bibitem[Zhu et~al.(2025)Zhu, Zhang, Shao, and Tang]{zhu2025kv}
Tianrui Zhu, Shiyi Zhang, Jiawei Shao, and Yansong Tang.
\newblock Kv-edit: Training-free image editing for precise background preservation.
\newblock \emph{arXiv preprint arXiv:2502.17363}, 2025.

\end{thebibliography}

}
\end{document}